\documentclass[12pt,oneside,english]{amsart}
\usepackage[T1]{fontenc}
\usepackage[utf8]{inputenc}
\usepackage{geometry}
\usepackage{fancyhdr}
\usepackage{color}
\usepackage{amstext}
\usepackage{amsthm}
\usepackage{amssymb}
\usepackage{mathdots}
\usepackage{esint}
\usepackage[foot]{amsaddr}
\usepackage{subcaption}

\makeatletter
\numberwithin{equation}{section}
\numberwithin{figure}{section}
\theoremstyle{plain}
\newtheorem{thm}{\protect\theoremname}
  \theoremstyle{plain}
  
  \theoremstyle{remark}
  \newtheorem*{rem*}{\protect\remarkname}
  \theoremstyle{remark}
  
  \theoremstyle{plain}
  
  \theoremstyle{plain}
  \newtheorem{exa}[thm]{\protect\examplename}

\usepackage[english]{babel}
\usepackage{fancyhdr}
\providecommand{\lemmaname}{Lemma}

\providecommand{\theoremname}{Theorem}
\providecommand{\corollaryname}{Corollary}
\providecommand{\examplename}{Example}

\newcommand{\abs}[1]{\ensuremath{|#1|}}

\newcommand{\cF}{\mathcal{F}}

\newcommand{\cO}{\mathcal{O}}

\DeclareMathOperator*{\argmin}{arg\,min}

\usepackage{babel}
\providecommand{\lemmaname}{Lemma}
  
  \providecommand{\remarkname}{Remark}
\providecommand{\theoremname}{Theorem}

\usepackage{babel}
\providecommand{\lemmaname}{Lemma}
  
\providecommand{\theoremname}{Theorem}

\usepackage{babel}
\providecommand{\theoremname}{Theorem}

\usepackage{babel}
\providecommand{\theoremname}{Theorem}

\usepackage{mathtools}

\DeclarePairedDelimiterX{\inp}[2]{\langle}{\rangle}{#1, #2}

\makeatother

\usepackage{babel}
  \providecommand{\corollaryname}{Corollary}
  \providecommand{\lemmaname}{Lemma}
  \providecommand{\remarkname}{Remark}
\providecommand{\theoremname}{Theorem}

\begin{document}

\title{Hedging of Financial Derivative Contracts\\ via Monte Carlo Tree Search}

\author{Oleg Szehr}
\address[Oleg Szehr]{Dalle Molle Institute for Artificial Intelligence (IDSIA) - SUPSI/USI, Manno, Switzerland}
\email{oleg.szehr@idsia.ch}

\begin{abstract}
The construction of approximate replication strategies for pricing and hedging of derivative contracts in incomplete markets is a key problem of financial engineering. Recently Reinforcement Learning algorithms for hedging under realistic market conditions have attracted significant interest. While research in the derivatives area mostly focused on variations of $Q$-learning, in artificial intelligence Monte Carlo Tree Search is the recognized state-of-the-art method for various planning problems, such as the games of Hex, Chess, Go,... This article introduces Monte Carlo Tree Search as a method to solve the stochastic optimal control problem behind the pricing and hedging tasks. As compared to $Q$-learning it combines Reinforcement Learning with tree search techniques. As a consequence Monte Carlo Tree Search has higher sample efficiency, is less prone to over-fitting to specific market models and generally learns stronger policies faster. In our experiments we find that Monte Carlo Tree Search, being the world-champion in games like Chess and Go, is easily capable of maximizing the utility of investor's terminal wealth without setting up an auxiliary mathematical framework.
\end{abstract}

\maketitle

\section{\label{subsec:itro}Introduction }
Monte Carlo Tree Search (MCTS) is an algorithm for approximating optimal decisions in multi-period optimization tasks by
taking random samples of actions and constructing a search tree according to the results. In artificial intelligence MCTS is the state-of-the-art and most well-researched technique for solving sequential decisions problems in domains that can be represented by decision trees. Important applications include the search for optimal actions in planning problems and games with perfect information. This article introduces the application of MCTS to an important stochastic planning problem in Finance: The pricing and hedging of financial derivative contracts under realistic market conditions.

The modern derivatives pricing theory came to light with two articles by Black and Scholes~\cite{BS1973} and Merton~\cite{merton1973} on the valuation of~\emph{option contracts}. The articles introduce what is nowadays known as the Black-Scholes-Merton (BSM) economy: A minimal model of a financial market comprised of a risk-free asset (commonly called a 'bond' or 'cash'), a risky asset (the 'stock') and an asset, whose price is derived (thus a~\emph{derivative}) from the stock. The risk-free asset exhibits a constant rate of interest, the risky asset promises higher returns but bears market fluctuations (following a continuous-time Geometric Brownian motion (GBM) process), the derivative contract, a so-called European call option, gives its holder the right to purchase shares of stock at a fixed price and date in the future. BSM show that in their economy there exists a self-financing, continuous-time trading strategy in the risk-free asset and the stock that exactly replicates the value (i.e.~price) of the option contract over the investment horizon. The replicating portfolio can be used to offset (to 'hedge') the risk involved in the option contract. The absence of arbitrage then dictates that the option
price must be equal to the cost of setting up the replicating/hedging portfolio. 

In reality, continuous time trading is of course impossible. It cannot even serve as an approximation due to high resulting transaction costs. In fact transaction costs can make it cheaper to hold a portfolio of greater value than the derivative (i.e.~to super-hedge)~\cite{Bensaid1992}. In other words, the transaction costs create~\emph{market incompleteness}:~the derivative contract cannot be hedged exactly and, as a consequence, its issuer incurs~\emph{risk}. Further reasons for market incompleteness include the presence of mixed jump-diffusion price processes~\cite{Aase1988} and stochastic volatility~\cite{HULL1987}. Since hedging under market-incompleteness involves risky positions it follows that the price depends on the risk preference of the investor. Hence, in reality, the hedging problem is 'subjective', it depends on the investors~\emph{utility}.

In modern terms the replication and pricing problems are often phrased as stochastic multi-period utility optimization problems~\cite{duffie2001}: In order to hedge a derivative contract an economic agent consecutively purchases/sells shares of stock as to maximize a given utility function of her/his terminal holdings. Since many years dynamic programming-based algorithms constitute a standard tool for utility optimization in the incomplete market setting, see e.g.~\cite{hodges1989option,Barron1990,ElKaroui1995,duffie2001,Zakamouline2006} and references therein. More recently reinforcement learning (RL) has gained wide public attention, as RL agents, trained tabula-rasa and solely through self-play, reached super-human level in Atari~\cite{mnih2013playing} and board games such as Hex~\cite{Anthony2017}, Go~\cite{silver2017mastering} and Chess~\cite{Silver2018}. This advancement has also reached the financial derivatives area, where several publications reported on promising hedging performance of trained RL agents, see e.g.~\cite{halperin2017qlbs,kolm2019dynamic,cao2019deep,Bisi2020,Vittori2020}. The article \cite{halperin2017qlbs} proposes Deep $Q$-Learning (DQN) \cite{watkins1992q, osband2016deep} to address utility optimization with a quadratic utility functional but with no transaction costs. Recent work \cite{cao2019deep} applies double $Q$-learning \cite{van2016deep} to take account of stochastic volatility. On the technical side the core of~\cite{halperin2017qlbs,cao2019deep} lies in the decomposition of (a quadratic) terminal utility into 'reward portions' that are granted to a DQN agent. The articles~\cite{buehler2019deep,buehler2019deep2} combine Monte Carlo simulation with supervised learning for policy search. The article~\cite{kolm2019dynamic} assumes a 'semi-supervised' setting, where derivative prices are known a priori and DQN is employed to trade off transaction costs versus replication error for {{mean-variance equivalent}} loss distributions. Article~\cite{Vittori2020} studies risk-averse policy search for hedging assuming, too, that price information is provided a priori and following the mean-volatility approach of~\cite{Bisi2020}.
 
The article at hand introduces the application of MCTS to solve the planning problems of pricing and hedging in a setting where no a priori pricing information is given. Despite the focus on games in artificial intelligence research, MCTS applies to Markov decision processes whenever the planning problem can be modeled in terms of a decision tree. This is apparent already in the early roots of MCTS~\cite{Kearns2002, Kocsis2006}, where Monte Carlo search has been investigated as a method to solve stochastic optimal control problems. Over the years MCTS has been applied to various planning problems, including single-player games (puzzles)~\cite{Schadd} and the travelling salesman problem~\cite{rimmel}. The conceptually closest applications of MCTS to our hedging problem are the planning problems for energy markets in~\cite{CouetouxPhD} and the stochastic shortest path problems commonly summarized within the~\emph{sailing domain}~\cite{vanderbei1996} paradigm. As compared to un-directed Monte Carlo methods (including DQN), MCTS is characterized by problem-specific and heavily restricted search. As a consequence MCTS learns stronger policies faster. It is also less susceptible to the typical instability of RL methods when used in conjunction with deep neural networks~\cite{tsitsiklis1997analysis}. See~\cite{Kocsis2006} for a comparison of MCTS versus plain RL agents in the sailing domain or~\cite{Anthony2017} in the game of Hex. 

Stability and sample efficiency are key when it comes to the application of RL-based algorithms in practice. The relative sample inefficiency of RL as compared to the supervised learning-based methods~\cite{buehler2019deep,buehler2019deep2} leaves the practitioner little choice but to train on simulated data, as e.g.~in~\cite{buehler2019deep2,kolm2019dynamic,cao2019deep}. However, an agent trained in a simulated environment learns to acquire rewards in~{this} environment and not necessarily in the real financial markets. MCTS combines RL with~{search}, which leads to a stronger out-of-sample performance and less over-fitting to a specific market model. In a nutshell the advantage of MCTS-based hedging as compared to other methods can be summarized as follows:
\begin{enumerate}
\item MCTS directly approaches the multi-period optimal control problem of maximizing the utility of terminal wealth. Returns are only granted at contract maturity.
\item MCTS operates under any model for the underlying market, transaction costs, utility, price, etc.
\item As compared to other RL methods, MCTS learns stronger actions faster.
\item MCTS combines RL with search, which implies a stronger generalization ability, in particular when trained on a simulated market environment.
\end{enumerate}

\begin{figure}
\centering
\begin{subfigure}[b]{0.45\textwidth}
\includegraphics[width=\linewidth]{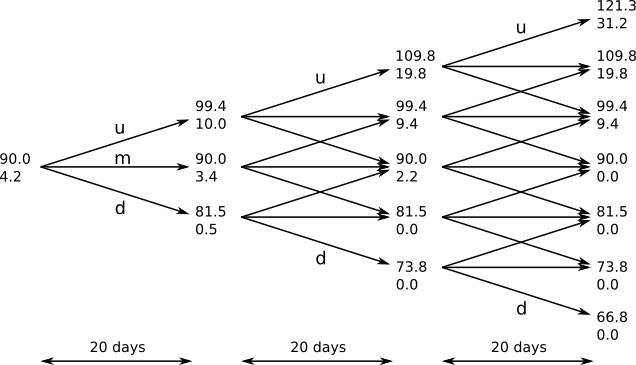}
\caption{State transitions in trinomial market.}
\end{subfigure}
\begin{subfigure}[b]{0.45\textwidth}
\includegraphics[width=\linewidth]{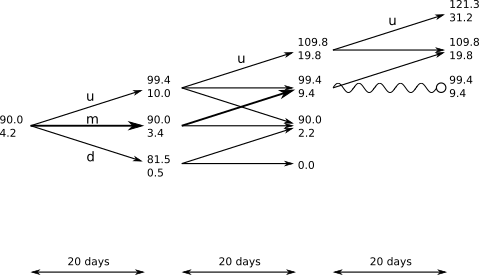}
\caption{Asymmetric planning in MCTS.}
\end{subfigure}
\caption{Illustration of partial planning trees for valuation of a European call option in trinomial and MCTS models. The initial stock and strike prices are $90$, the planning horizon is $60$ days, computed in $3$ time steps. The volatility of the underlying stock model is $30\%$. No interest and dividends are paid. Stock prices are shown in the first row at each node, option prices in the second. The trees depict the evolution of the market (the agent's actions are not shown). The MCTS planning tree is heavily restricted and asymmetric leading to a more efficient and realistic search. Tree constuction is guided by a dedicated tree policy. Leaf nodes are valued by Monte Carlo rollout (wavy line).}\label{trinomialFigure}
\end{figure}

Finally let us remark that the application of tree-search techniques to hedging is not a surprise. The binomial~\cite{Cox1979} and trinomial~\cite{Boyle1986} tree models are popular discrete approximation methods for the valuation of option contracts, with and without stochastic volatility. Beginning with the original work~\cite{hodges1989option} binomial trees have been employed as a tool for utility optimization and computation of reservation price. The main idea behind the application of MCTS is to reduce the size of the planning tree with emphasis on relevant market moves and actions. In our applications we will only address the discrete time, discrete market, discrete actions setting. For completeness we mention that MCTS has been studied also in the setting of continuous action and state spaces~\cite{CouetouxPhD,kim2020}. MCTS can be employed to simulate both agent and market behavior. To illustrate our idea, Figure~\ref{trinomialFigure} shows a comparison between a trinomial market model and an MCTS market model for the pricing of an European call option. The trinomial model is characterized by a discrete stock market, where at each time step a transition $up, middle, down$ leads from $S_t$ to
$S_{t+1}\in\{u S_t,mS_t,dS_t\}$. MCTS builds a restricted and asymmetric decision tree, see Section~\ref{hedgingAsMDP} for details.
\section{The Hedging and Pricing Problems}
\label{hedging}
We consider an economy comprised of cash, shares of risky asset and a derivative contract. Let $S_t$ denote the price of the underlying stock at time $t$, which is assumed to follow a Markov process. New market information becomes available at discrete times $t=1,2,...,T$. At time $t=0$ the investor sells the derivative contract and subsequently maintains a replication portfolio of the form
$$\Pi_t=n_tS_t+B_t$$
with $n_t$ shares of the underlying and a bank account\footnote{For simpler notation we assume that there is no interest on cash.} $B_t$. Shortly before new market information becomes available, the investor decides upon the new portfolio composition. The agent's objective is to maximize the expected utility 
$$\max \mathbb{E}[u(w_T)]$$
of wealth over the investment horizon $T$. The terminal wealth $w_T$ is comprised of the value of the replication portfolio $\Pi_T$, the negative payoff of the derivative contract at maturity as well as all costs acquired over the investment horizon. The utility function $u$ is assumed to be smooth, increasing and concave. In what follows we describe the pricing and hedging problem for the particular case of an European vanilla call option. The latter gives its holder the right to purchase, at maturity $T$, shares of the underlying at price $K$, i.e.~its terminal value is
$$C_T=C_T(S_T)=(S_T-K)^+.$$
Notice that the same discussion applies to arbitrary payoff structures. We refer to~\cite{duffie2001} for a detailed introduction to dynamic asset allocation for portfolio optimization and hedging.

\subsection{The complete market}\label{hedgingCompleteMarket}
The complete market is characterized by the absence of transaction costs and  the assumption that all derivatives can be replicated exactly. The composition of the replication portfolio is computed by back-propagation imposing that the trading strategy is {self-financing}. The latter asks that shares bought at time $t$ are equally billed to the bank account. In the absence of transaction costs this implies
\begin{align}
\Pi_{t+1}-\Pi_t=n_{t+1}S_{t+1}+B_{t+1}-n_tS_t-B_t=n_{t+1}\Delta S_{t+1},\label{selfFinancing}
\end{align}
where we use the notation $\Delta X_{t+1}=X_{t+1}-X_t$ throughout the article. Consequently, when starting with an initial portfolio value of $\Pi_0$ (from the sell of the derivative contract) we have
\begin{align*}
\Pi_{T}=\Pi_0+\sum_{t=0}^{T-1}n_{t+1}\Delta S_{t+1}.
\end{align*}
Back-propagation means that the composition of the replication portfolio is computed iteratively beginning at maturity. Suppose, for the moment, that the investor has issued the option at $t=0$ and, in return, he/she received an amount of $C_0$ in cash. In a complete market the choice of $C_0$ is 'trivial' in the sense that in the absence of arbitrage there exists a deflator process that induces a unique martingale measure and price. Let us define the optimal value function by
\begin{align*}
V^*(t,S_t,\Pi_t):=\max_{n_1,...,n_{T}}\mathbb{E}[u(\Pi_T-C_T)|\cF_t].
\end{align*}
The expression $\mathbb{E}[\cdot|\cF_t]$ denotes a conditional expectation with respect to the (natural)~\emph{filtration} $\cF_t$ (of $S_t$). Conditioning on $\cF_t$ means that all market information at time $t$, including $S_t$, $\Pi_t$,..., is realized and available for decision making. We give a more detailed discussion of $\cF_t$ in terms of a decision tree in Section~\ref{decisionTree}. The maximum is taken over the predictable sequences of investment decisions. Concretely, the dynamic programming formulation of the hedging exercise is as follows. One step before maturity the agent is faced with finding the portfolio composition $\Pi_{T}$ such as to maximize
\begin{align*}
\mathbb{E}[u(C_T-\Pi_T)|\cF_{T-1}].
\end{align*}
Taking account of the self-financing constraint~\eqref{selfFinancing} this is equivalent to computing the optimal value function at $T-1$ via
\begin{align*}
\max_{n_{T}}\mathbb{E}[u(C_T-\Pi_{T-1}-n_T\Delta S_{T})|\cF_{T-1}]=V^*(T-1,S_{T-1},\Pi_{T-1}).
\end{align*}
Two steps before maturity the agent is faced with the choice of $n_{T-1}$ to maximize 
\begin{align*}
\max_{n_{T-1}}\mathbb{E}[V^*(T-1,S_{T-1},\Pi_{T-1})|\cF_{T-2}]=V^*(T-2,S_{T-2},\Pi_{T-2}).
\end{align*}
More generally, no matter what the previous decisions have been, at time $t$ the optimal number of shares is obtained from
\begin{align}
\max_{n_{t+1}}\mathbb{E}[V^*(t+1,S_{t+1},\Pi_{t+1})|\cF_t]=V^*(t,S_{t},\Pi_{t}).\label{hedgeBellman}
\end{align}
As described in Section~\ref{MDPs} this is a form of the Bellman equation.
\begin{exa}\label{BSM}
\emph{Black-Scholes-Merton (BSM) least variance hedges~\cite{merton1990}:} In the BSM economy the market is comprised of the three assets $S_t,B_t,C_t$ and individual wealth increments are $\Delta w_{t+1} =  n_{t+1}\Delta S_{t+1}-\Delta C_{t+1}$. BSM pricing and hedging rely on the paradigm that the hedging portfolio should exactly replicate the option contract at all times. Accordingly, it is assumed that utility is additive with respect to individual wealth increments,
$$ \mathbb{E}[u(w_T)] = u_0(w_0)+\sum_{t=0}^{T-1}\mathbb{E}[u_{t+1}(\Delta w_{t+1})].$$
The individual
utility increments are given by the variances
\begin{align}
\mathbb{E}[u_{t+1}(\Delta w_{t+1})|\cF_t] = -\mathbb{V}\left[\Delta C_{t+1} - n_{t+1}\Delta S_{t+1}\ |\cF_t \right].\label{BSReward}
\end{align}
For this form of utility the investor chooses the portfolio composition at any time step such as to minimize the variance (risk) of the consecutive investment period. Assuming $S_t$ follows a geometric Brownian motion, the well-known solution is the Black-Scholes $\delta$-hedge portfolio. In the limit of small time steps and in absence of transaction costs the best hedges are given by $ n_{t+1}\rightarrow\delta_t = \partial C_t/\partial S_t$.  In this case the global minimum of the utility is reached: In following the $\delta$-hedge strategy the agent does not take any risk, any deviation from this strategy results in a risky position.
\end{exa}
\subsection{Incomplete markets and pricing}
In the presence of transaction costs equation~\eqref{selfFinancing} must be modified, where each trading activity is accompanied with a loss of $transactionCost<0$. In other words
\begin{align}
\Pi_{t+1}-\Pi_t=n_{t+1}\Delta S_{t+1}+transactionCosts.\label{constraintselfFinTransactionCosts}
\end{align}
An additional liquidation fee might apply at maturity leading to a terminal payoff of $C_T - \Pi_T +liquidationCost$. The dynamic programming approach outlined in~Section~\ref{hedgingCompleteMarket} remains valid if proper account is taken of the payoff structure and the new self-financing requirement~\eqref{constraintselfFinTransactionCosts}. Notice that the structure of the hedging problem might require additional state variables. If, for instance, transaction costs depend explicitly on the current holdings $n_t$ then so does the value function $V(n_t,S_t,\Pi_t)$. In similar vein stochastic volatility and interest models might require additional state variables to keep~track of holdings in stock and cash. 
Notice also that in the context of incomplete markets the choice of price has to be included as part of the optimization problem. Various models have been proposed to address pricing and hedging in incomplete markets. One prominent line that was initiated by Föllmer and Sondermann~\cite{Foll1985}, see also~\cite{Bouchaud1994,Schl1994, Schweizer1995}, considers pricing through a risk minimization procedure. The 'fair hedging price'~\cite{Schl1994} is the minimizer of the 'risk'
\begin{align*}
F_0 = \argmin_{\Pi_0}\left[\min_{n_1,...,n_{T}}\mathbb{E}[-u(\Pi_T-C_T)]\right].
\end{align*}
We illustrate this by the standard example.
\begin{exa}\label{terminal}
\emph{Terminal variance hedging~\cite{Schweizer1995}:} The problem is to minimize the expected net quadratic loss
\begin{align*}
\min_{\Pi_0,n_1,...,n_{T}}\mathbb{E}[(\Pi_T-C_T)^2]
\end{align*}
simultaneously over the cash balance $\Pi_0$ from the sell and the hedging decisions $n_1$,...,$n_T$.
As in Section~\ref{hedgingCompleteMarket} dynamic programming applies to this problem, where the value function involves an additional optimization over $\Pi_0$,
\begin{align*}
V^*(t,S_t,\Pi_t)=\min_{\Pi_0,n_1,...,n_{T}}\mathbb{E}[(\Pi_T-C_T)^2|\cF_t].
\end{align*}
Notice that if this is solved by $\Pi_0^*$ and investment decisions $n_1(\Pi_0^*),...,n_{T}(\Pi_0^*)$, then these decisions are also optimal in terms of the minimization of the variance of terminal wealth, i.e.
\begin{align*}
\min_{n_1,...,n_{T}}\mathbb{V}[C_T-\Pi_T].
\end{align*}

\end{exa}
This form of pricing ignores the possibility that a portfolio without liabilities might have a positive value. The main idea behind the concept of reservation price~\cite{hodges1989option,Davis1993,Clewlow1997,Andersen1999,Zakamouline2006}  is that buy/sell prices of derivative contracts should be such that the buyer/seller remains indifferent, in terms of expected utility, with respect to the two situations:
\begin{enumerate}
\item Buying/ Selling a given number of the derivative contracts and hedging the resulting risk by a portfolio of existing assets, versus
\item leaving the wealth optimally invested within existing assets (and not entering new contracts).
\end{enumerate}
Suppose an investor sells $\theta>0$ options at time $t$ and is not allowed to trade the options thereafter. The seller's problem is to choose a trading strategy (policy, see below) subject to~\eqref{constraintselfFinTransactionCosts} to maximize expected utility of the resulting portfolio
\begin{align*}
{V}_{sell}^*(t,n_t,S_t,\Pi_t)=\max_{}\mathbb{E}[u(\Pi_T-\theta C_T)|\cF_t].
\end{align*}
In the absence of options $\theta=0$, the investor simply optimizes
\begin{align*}
\tilde{V}^*(t,n_t,S_t,\Pi_t)=\max_{}\mathbb{E}[u(\Pi_T)|\cF_t].
\end{align*}
The reservation sell price of $\theta$ European call options is defined as the price $P_\theta^s$ such that
\begin{align*}
\tilde{V}^*(t,n_t,S_t,\Pi_t)={V}_{sell}^*(t,n_t,S_t,\Pi_t+\theta P^s_\theta).
\end{align*}
The buy price is defined by analogy. It is well-known that in the absence of market friction the buy and sell prices converge to the Black-Scholes price. The computation of reservation price is simplified if the investor's utility function exhibits
a constant absolute risk aversion (a so-called CARA investor). For the CARA investor the option price and hedging strategy are independent of the investor's total wealth. In case of proportional transaction costs the problem can be solved explicitly.

\begin{exa}\label{hodgesExample}
\emph{Hodges-Neuberger (HN) exponential utility hedges~\cite{hodges1989option,Davis1993}:}  Assuming a constant proportional cost model,
\begin{align*}
transactionCosts(\Delta n_{t+1},S_t)=-\beta\abs{\Delta n_{t+1}}S_t,
\end{align*}
and exponential utility of terminal wealth (CARA),
$$
u(w_T)=-\exp(-\lambda w_T),
$$
results in trading strategies that are wealth-independent and that do not create risky positions when no derivative contract is traded. The portfolio allocation space is divided into three regions, which can be
described as the Buy region, the Sell region, and the no-transaction
region. If a portfolio lies in the Buy/ Sell region, the optimal strategy is to buy/ sell
the risky asset until the portfolio reaches the boundary between the Buy/ Sell
region and the no-transaction region. If a portfolio is located in the no-transaction region then it is not adjusted at the respective time step.
%
\end{exa}
\section{Monte Carlo Planning}
\subsection{Markov Decision Processes}\label{MDPs}
Markov decision processes (MDPs) provide a mathematical framework
for modeling sequential decision problems in uncertain dynamic environments~\cite{Puterman1994}. An MDP is a four-tuple $(S, A, P, R)$, where $S=\{s_1,...,s_n\}$ is a set
of states, $A=\{a_1,...,a_m\}$ is a set of actions\footnote{We assume that $S$ and $A$ are finite in what follows.}. The transition probability function $P:S \times A\times S\rightarrow [0,1]$ associates the probability of entering the next state $s'$ to the triple $(s',a,s)$. The reward obtained from action $a$ in state
$s$ is $R(s, a)$. The policy $\pi :S\times A\rightarrow [0,1]$ is the conditional probability of choosing action $a$ in state $s$. The value function of policy $\pi$ associates to each state $s$ the expected total reward when starting at
state $s$ and following policy $\pi$,
\begin{align*}
V^\pi(s)=\mathbb{E}_{\pi}\left[\sum_{t=0}^T\gamma^tR_t(s_t,a_t)\Big |s_0=s\right].
\end{align*}
The role of the discount factor $\gamma\in[0,1]$ is two-fold. In the case of infinite time horizons $\gamma<1$ is chosen to discount the value of future rewards and to enforce convergence of the above series\footnote{We assume $T<\infty$ in what follows.}. Second, $\gamma$ measures how relevant in terms of total reward are the immediate consequences of an action versus its long-term effect. The agent's goal is to find a policy that maximizes the expected total reward. The optimal value function
$$V^*(s)=V^{\pi^*}(s)=\sup_\pi V^\pi(s)$$
satisfies the Bellman equation
\begin{align}
V_t^*(s)=\max_a\left\{R_t(s,a)+\gamma\sum_{s'\in S}P(s'|s,a)V_{t+1}^*(s')\right\}.\label{BellmanFixedPoint}
\end{align}

\subsection{The $k$-armed bandit problem}\label{bandits}
The $k$-armed bandit is the prototypical instance of RL reflecting the simplest MDP setting~\cite{sutton2018reinforcement}. Formally the setup involves a set of $k$ possible actions, called arms\footnote{The name originates from a gambler who chooses from $k$ slot machines.}, and a sequence of $T$ periods. In each period $t = 1,...,T$ the agent chooses an arm $a\in\{a_1,..,a_k\}$ and receives a random reward $R=X_{i,t}$, $1\leq i\leq k$. The main assumption is that the random variables $X_{i,t}$ are independent with respect to $i$ and i.i.d.~with respect to $t$. As compared to general MDPs the $k$-armed bandit addresses a simplified scenario, where the current action determines the immediate reward but has no influence on subsequent reward. This leads to the exploitation-exploration dilemma: should the agent choose an action that has been lucrative so far or try other actions in hope to find something even better. For large classes of reward distributions, there is no policy whose regret after $n$ rounds grows slower than $\cO(\log n)$~\cite{Lai1985}. The UCB1 policy of~\cite{auer2002finite} has expected regret that achieves the asymptotic growth of the lower bound (assuming the reward distribution has compact support). UCB1 keeps track of the average realized rewards $\bar X_i$ and selects the arm that maximizes the confidence bound:
$$UCB1_i = \bar{X}_i+wc_{n,n_i}\ \textnormal{with}\ c_{n,l}=\sqrt{\frac{2\ln(n)}{l}},$$
where $n_i$ is the number of times arm $i$ has been played so far. The average reward $\bar{X}_i$ emphasizes exploitation of the currently best action, while the $c_{n,n_i}$ encourages the exploration of high-variance actions. Their relative weight is determined by a problem-specific hyper-parameter $w$.

\subsection{Algorithms for Markov Decision Processes}\label{algosMDPs}

For complex MDPs, the computation of optimal policies is usually intractable. Several approaches have been developed to compute near optimal policies by means of function approximation and simulation. 

The value iteration algorithm calculates an approximation to the optimal policy by repeatedly
performing Bellman’s updates over the entire state space:
\begin{align*}
V(s)\leftarrow\arg\max_{a\in A(s)}\left\{R(s,a)+\gamma\sum_{s'\in S}P(s,a,s')V(s')\right\},\ \forall s\in S,
\end{align*}
where $A(s)$ denotes the set of applicable actions in state $s$. Value iteration converges to the optimal value function $V^*$. The drawback of this method is that it performs updates over the entire state space. To focus computations on relevant states, real-time heuristic search methods have been generalized to non-deterministic problems~\cite{Dean1995,Hansen2001}. However these algorithms only apply in situations, where the transition probabilities $P(s'|s,a)$ are known and the number of possible successors states for a state/action pair remains low. If $P(s'|s,a)$ are not given it is usually assumed that an {environment simulator} is available that generates samples $s'$ given $(s,a)$ according to $P(s'|s,a)$. This has been proposed for example in~\cite{Kearns2002},
where a state $s$ is evaluated by trying every possible action $C$ times and, recursively, from each generated state every possible action $C$ times, too, until the planning horizon $H$ is reached.
While theoretical results~\cite{Kearns2002} demonstrate that such search provides near optimal policies for any MDP, often $H$ and $C$ need be so large that computation becomes impractical. The key idea of~\cite{Chang2005, Kocsis2006, Kocsis2006_} is to incrementally build a problem-specific, restricted and heavily asymmetric decision tree instead of 'brute force Monte Carlo' with width $C$ and depth $H$. The growth of the decision tree is controlled by a \emph{tree policy}, whose purpose it is to efficiently trade-off the incorporation of new nodes versus the simulation of existing promising lines. In each iteration, more accurate sampling results become available through the growing decision tree. In turn they are used to improve the tree policy. Algorithms of this form are commonly summarized under the name Monte Carlo Tree Search (MCTS)~\cite{Browne2014}. The MCTS' main loop contains the following steps:
\begin{enumerate}
\item \emph{Selection:} Starting at the root node, the tree policy is applied to descend through
the tree until the most relevant expandable node is
reached.
\item \emph{Expansion:} Child nodes are added to
expand the tree, again according to tree policy.
\item \emph{Simulation:} A simulation is run from the new nodes.
\item \emph{Backpropagation:} The simulation result is used to update the tree policy.
\end{enumerate}
Kocsis and Szepesvári~\cite{Kocsis2006,Kocsis2006_} propose to follow UCB1 bandit policy for tree construction. They demonstrate that (in case of deterministic environments) the UCB1 regret bound still holds in the non-stationary case and that given infinite computational resources the resulting MCTS algorithm, called UCT, selects the optimal action. In summary UCT is characterized by the following features, which make it a suitable choice for real-world applications such as hedging: 
\begin{enumerate}
\item UCT is aheuristic. No domain-specific knowledge (such as an evaluation for leaf-nodes) is needed.
\item UCT is an anytime algorithm.
Simulation outcomes are back-propagated immediately, which ensures that tree statistics are up to date after each iteration. This leads to a
small error probability if the algorithm is stopped prematurely.
\item UCT convergence
to the best action if enough resources are granted.
\end{enumerate}
In the setting of adversarial games the role of the simulator is usually taken by another instance of the same MCTS algorithm. The success of MCTS methods, especially in games such as Hex~\cite{Anthony2017}, Go~\cite{silver2017mastering} and Chess~\cite{Silver2018}, is largely due to the mutual improvement of policies for tree construction and node evaluation. Various architectures have been proposed for the 'search and reinforce' steps. Our specific architecture is described in Section~\ref{ourArchitecture}.

\section{Monte Carlo Tree Search for Hedging}

\subsection{Hedging as a Markov Decision Process}
\label{hedgingAsMDP}
Two formulations of the hedging exercise are present in the literature. In the P\& L formulation, see e.g.~\cite{kolm2019dynamic,Vittori2020}, it is assumed that at each time the investor is aware of the price of the derivative contract. The price might be available through an ad hoc formula or learned from the market. Typically the investor will set up a hedging portfolio to match the price of the derivative contract or to balance between replication error and transaction costs. Notice that if market incompleteness is introduced 'on top' of a pricing model for complete markets, the respective prices cease to be valid~{even in an approximate sense}. The Cash Flow formulation addresses the full-fledged utility optimization problem of Section~\ref{hedging} without pricing information. If the derivative contract has a unique Cash Flow at maturity then the Belmann Equation~\eqref{BellmanFixedPoint} takes the form~\eqref{hedgeBellman}. Equation~\eqref{BellmanFixedPoint} reflects a derivative contract with a 'tenor structure' corresponding to a sequence of maturity times. To apply MCTS a simulator is required to sample from the probabilities\footnote{Remark that in standard microscopic market models in terms of It\^{o} diffusion processes, the agent's action does not influence the behavior of the market $P(s'|s,a)=P(s'|s)$.} $P(s'|s,a)$. Two models are possible:
\begin{enumerate}
\item \emph{Pure planning problem:} The probabilities $P(s'|s,a)$ are assumed a priori as a model of the market. They might be obtained by supervised learning from the real markets.
\item \emph{Adversarial game problem:} To focus search on relevant market behavior, the market is modeled by an independent market policy, which is adapted dynamically. This generalizes the scenario of the pure planning problem and resembles asymmetric games. The market policy can be learned as part of an independent instance of MCTS as in~Figure~\ref{trinomialFigure}.
\end{enumerate}
In our implementation we have focused only on setting (1), leaving (2) for further research. Figure~\ref{trinomialFigure} illustrates market actions in the setting (2).
\subsection{Building the decision tree}\label{decisionTree}

The planning tree consists of an alternating sequence of planned decisions and market moves. Depending on the underlying market model, the market either generates transitions from state $s$ to $s'$ with probability $P(s'|s,a)$ or a dynamic market policy guides the market behavior. Here we focus on the pure planning problem perspective, relying on a standard market model from the theory of stochastic processes.  The set $\Omega$ of all possible market states (over the planning horizon) is assumed to carry a filtration $\cF_t$, which models the unfolding of available information. $\cF_t$ can be interpreted\footnote{In fact $\cF_t$ is a sequence of $\sigma$-algebras and there exists a bijection that maps each $\sigma$-algebra to a partition of $S$.} as a sequence of partitions of $\Omega$. When trading begins no market randomness has been realized yet, $\cF_0 = \{\Omega\}$. At $T$ all market variables are realized, which corresponds to the partition of $\Omega$ into individual elements. At time $t$ the elements $P^{(t)}\in\cF_t=\{P_1,...,P_K\}$ reflect the realized state of the market. As the amount of available information increases the partitions of $\Omega$ become finer. The market evolution is represented by the nodes $(t,P^{(t)})$. 

\begin{figure}
\centering
\includegraphics[width=0.75\linewidth]{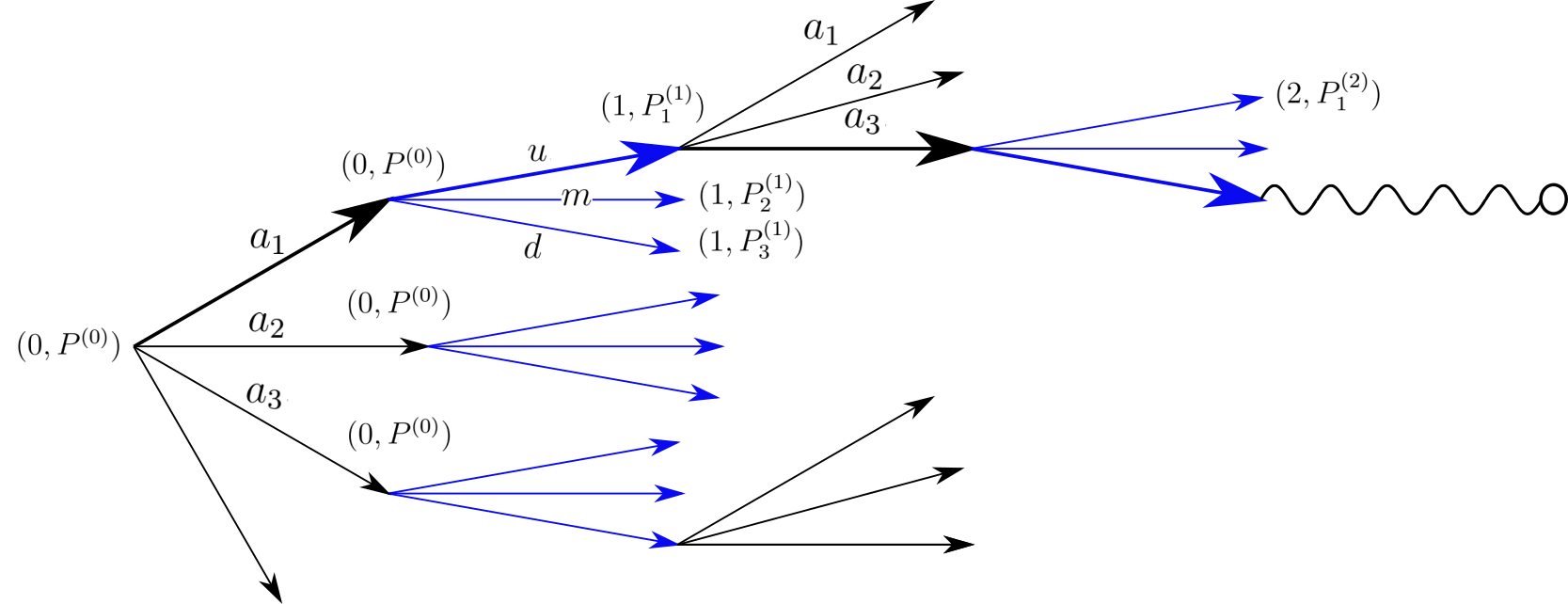}
\caption{Illustration of composed planning tree for a trinomial market model. Black arrow depict agent's actions. Blue arrows depict market moves. Nodes of tree are labeled by the elements $(t,P^{(t)})$ of the partition $\cF_t$ at time $t$. Currently searched path is highlighted. A Monte Carlo simulation is run to value the current leaf node (wavy line).}\label{planningTree}
\end{figure}

In our experiments we implemented a trinomial market model. The advantage of the trinomial model over the binomial
model is that drifts and volatilities of stocks can depend on the value of the underlying. Then there is
no dynamic trading strategy in the underlying security that replicates the
derivative contract resulting in risky positions for the hedger. Figure~\ref{planningTree} depicts the planning tree comprised of trinomial market model and MCTS guided actions. Notice that a planning tree structure comprised of agent and (random) market components has been proposed in~\cite{CouetouxPhD} for planning problems of energy markets, where MCTS in also employed in the continuous market setting.
\subsection{Architecture}\label{ourArchitecture} 
Our algorithm design orients itself at state-of-the-art MCTS for adversarial game tree search~\cite{Anthony2017,silver2017mastering}. In brief the UCT design is enhanced by making use of deep neural networks for policy and value function representation. Although the architectural details may vary, the main purpose of the involved neural networks is to
\begin{enumerate}
\item guide tree construction and search and to
\item provide accurate evaluation of leaf-nodes.
\end{enumerate}
Imitation learning is concerned with mimicking an expert policy $\pi^{E}$ that has been provided ad hoc. The expert delivers a list of strong actions given states and an apprentice policy $\pi^{A}$ is trained via supervised learning on this data. The role of the expert is taken by MCTS, while the apprentice is a standard convolutional neural network. The purpose of the expert is to accurately determine good actions. The purpose of the apprentice is to generalize the expert policy across possible states and to provide faster access to the expert policy. The quality of expert and apprentice estimates improve mutually in an iterative process. Specifically, regarding point~{(1)}, the apprentice policy is trained on tree-policy targets, i.e.~the average tree policy at the MCTS root reflects expert advice. The loss function is thus\footnote{Training the apprentice to imitate the optimal MCTS action $a^*$, $Loss_{Best}=-\log{\pi^{A}(a^*|s)}$, is also possible. However, the empirical results using $Loss_{TPT}$ are usually better.}
\begin{align*}
Loss_{TPT}=-\sum_{a}\frac{n_a}{n}\log{\pi^{A}(a|s)},
\end{align*}
where $n_a$ is the number of times action $a$ has been played from $s$ and $n$ is the total number of simulations. In turn the apprentice improves the expert by guiding tree search towards stronger actions. For this the standard UCB1 tree policy is enhanced with an extra term
\begin{align*}
NUCB1_a=UCB1_a+w_a\frac{\pi^A(a|s)}{n_a+1},
\end{align*}
where $w_a\sim\sqrt{n}$ is a hyper-parameter that weights the contributions of $UCB1$ and the neural network. Concerning point (2), it is well known that using good value networks substantially improves MCTS performance. The value network reduces search depth and avoids inaccurate rollout-based value estimation. As in the case of the policy network tree search provides the data samples $z$ for the value network, which is trained to minimize
\begin{align*}
Loss_{V}=-(z-V^A(s))^2.
\end{align*}
To regularize value prediction and accelerate tree search (1) and (2) are simultaneously covered by a multitask network with
separate outputs for the apprentice policy and value prediction. The loss for this network is simply the sum of $Loss_V$ and $Loss_{TPT}$.

\section{Experiments} 
\begin{figure}
\centering
\begin{subfigure}[b]{0.49\textwidth}
\includegraphics[width=\linewidth]{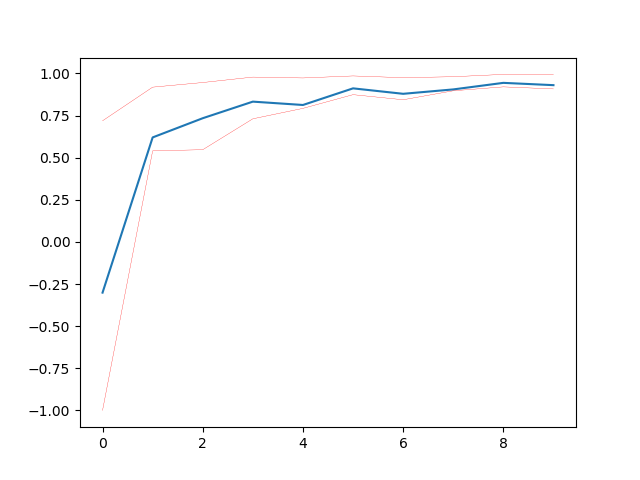}\caption{Terminal variance setting.}
\end{subfigure}
\begin{subfigure}[b]{0.49\textwidth}
\includegraphics[width=\linewidth]{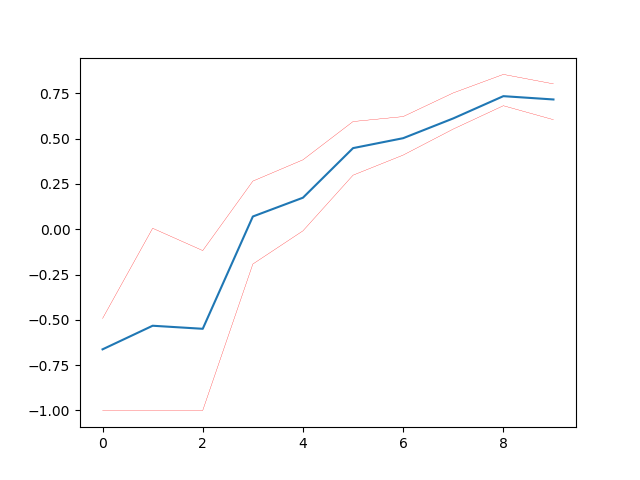}\caption{HN setting.}
\end{subfigure}
\begin{subfigure}[b]{0.49\textwidth}
\includegraphics[width=\linewidth]{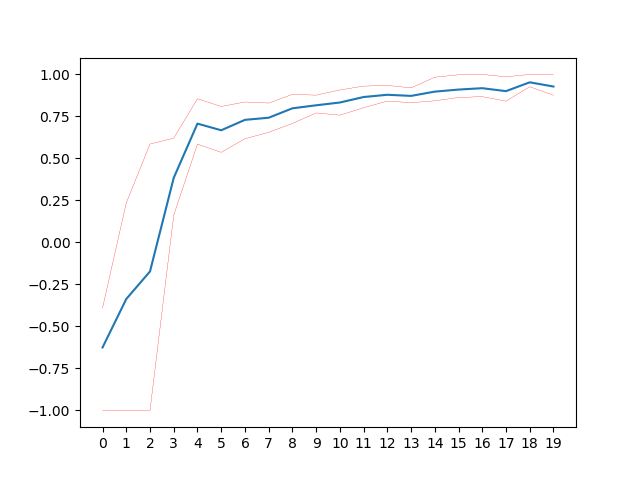}\caption{Training process in BSM setting. To reduce the effect of discretization error, the time to maturity interval has been split into $100$ (instead of $20$) steps.}
\end{subfigure}
\caption{Illustration of a training process of MCTS applied to terminal variance option hedging. Blue line depicts mean reward over $100$ random paths of market evolution, red lines depict the $25th$ and $75th$ percentiles.}\label{trainingProcess}
\end{figure}

\begin{figure}
\centering
\begin{subfigure}[b]{0.49\textwidth}
\includegraphics[width=\linewidth]{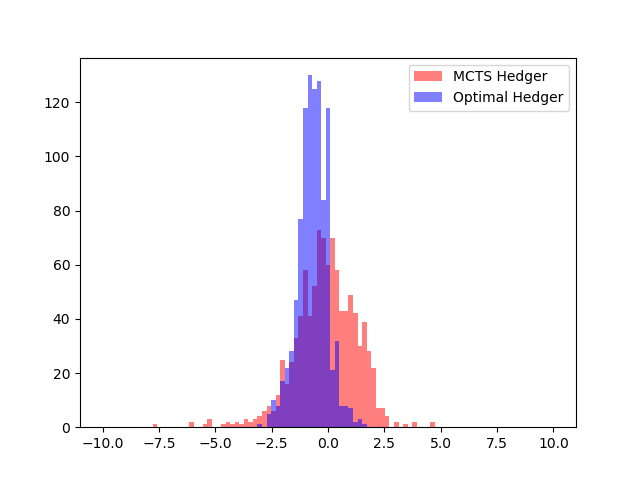}\caption{Terminal variance setting.}
\end{subfigure}
\begin{subfigure}[b]{0.49\textwidth}
\includegraphics[width=\linewidth]{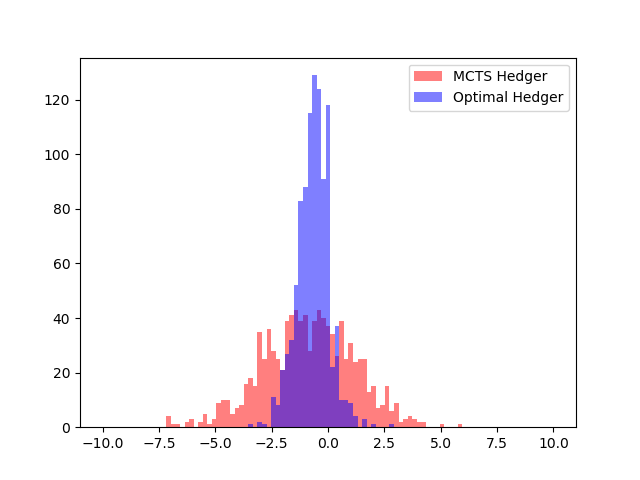}
\caption{BSM setting.}
\end{subfigure}
\caption{Histograms of terminal profit and loss of option and hedging portfolio at maturity as obtained on $1000$ random paths of market evolution. Agents have received $10$ training iterations each consisting of $1000$ episodes. For performance assessment the transaction costs are switched off. Red: trained MCTS agents. Blue: optimal agent.}\label{terminalPerformance}
\end{figure}

\begin{figure}
\centering
\begin{subfigure}[b]{0.49\textwidth}
\includegraphics[width=\linewidth]{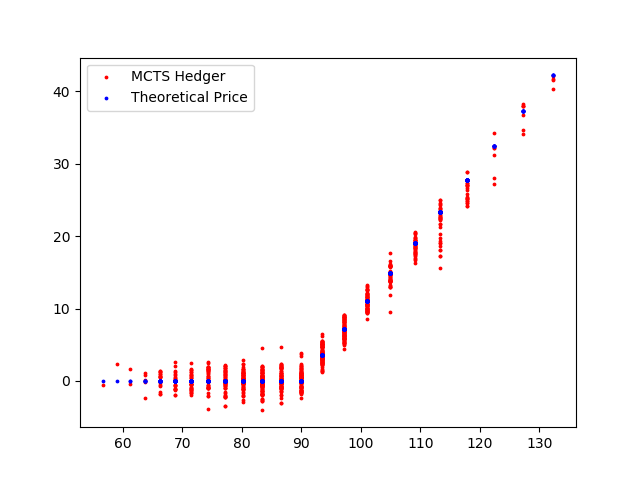}\caption{Terminal variance setting.}
\end{subfigure}
\begin{subfigure}[b]{0.49\textwidth}
\includegraphics[width=\linewidth]{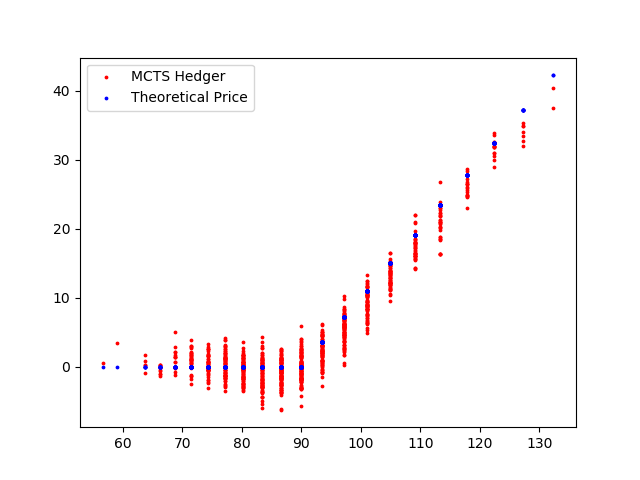}\caption{BSM setting.}
\end{subfigure}
\begin{subfigure}[b]{0.49\textwidth}
\includegraphics[width=\linewidth]{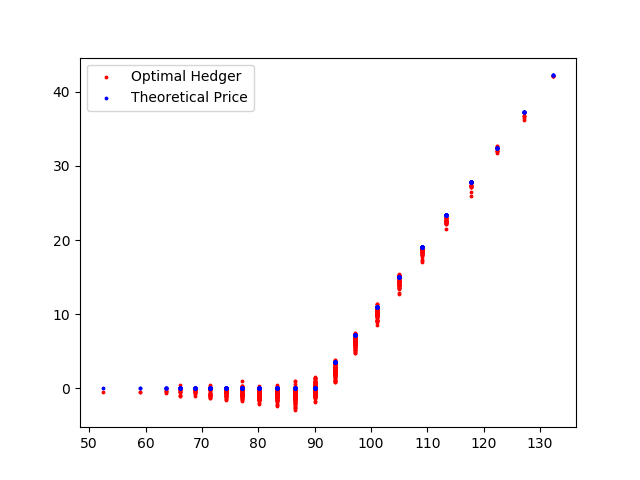}
\caption{Optimal hedger setting.}
\end{subfigure}
\caption{Scatter plots of terminal profit and loss of hedging portfolio at maturity as obtained on $1000$ random paths of market evolution. For performance measurement the transaction costs are switched off. To assess the level of discretization error (C) illustrates the performance of the optimal hedger. Red scatters: hedging agents. Blue scatters: theoretical option value at maturity.}\label{detailedTerminalPerformance}
\end{figure}

The code for conducting our experiments is available upon request for non-commercial use. Our architecture defines a general-purpose MCTS agent that, in principle, applies to any MDP-planning problem. This includes the valuation and hedging of American style and path-dependent derivative contracts in realistic markets. The purpose of our experiments is to provide proof-of-concept of MCTS hedging using toy examples of a small market and a European call option. The choice of the complex architecture is motivated by a potential application in an industry setting, where real-world market challenges ask for intelligent strategies. 
The following basic set of parameters was chosen for the experiments:
\begin{itemize}
\item \textbf{Option contract:} Short Euro vanilla call with strike $K=90$ and maturity $T=60$. Hedging and pricing are investigated under the reward models of Examples~\ref{BSM},~\ref{terminal},~\ref{hodgesExample}.
\item \textbf{Market:} Trinomial market model with initial price $S_0=90$ and constant stock volatility of $30\%$. No interest of dividends are paid. The planning horizon is split into $20$ discrete time steps.
\item \textbf{MCTS Parameters:} Each training iteration consists of $1000$ episodes, each of $25$ individual simulations. A discrete action space was chosen with $21$ possible action corresponding to the purchase or sell of up to $10$ shares or no transaction.
\item \textbf{MCTS Neural Network (NN):} States are represented as two-dimensional arrays. Present state symmetries are employed to enhance the training process. The neural network used for training consisted of $4$ CNN layers with ReLu activation and batch normalization, followed by dropout and linear layers.
\end{itemize}
In all hedging tasks rewards are episodic and gauged to the interval $[-1,1]$, where a reward of $1$ signifies a perfect hedge. After each iteration of training the out-of-sample performance is measured on $100$ trinomial tree paths. The trained and stored versions of the NN are compared in terms of their reward and the new version is accepted only if reward increases by a given percentage. Figure~\ref{trainingProcess} illustrates typical training processes in Examples~\ref{BSM},~\ref{terminal},~\ref{hodgesExample}. In each case the whole reward is granted at option maturity and no option prices are available during the training process. For the BSM example we accumulate the rewards over the training episode and increase the number of time steps to $100$ and the number of iterations to $20$ to investigate how close the trained agent gets to the BSM $\delta$-hedger. In the ideal BSM economy risk-less hedging is possible, which would yield a terminal reward of $1$. In contrast, in our setting discretization errors occur due to the finite representation of state and action spaces. Thus even in a complete market the optimal hedger, who is aware of the exact market model and option prices at each time step, could at best choose an action that minimizes hedging error in expectation. To assess the performance of the trained MCTS agents in the terminal variance and BSM settings we compare their terminal profits and losses versus the optimal hedger in a complete market. Figure~\ref{terminalPerformance} shows histograms of terminal profits and loss distributions obtained by executing trained hedger agents on $1000$ random paths of market evolution. To gain a clearer picture of the distribution of terminal profits and losses Figure~\ref{detailedTerminalPerformance} depicts a break down of terminal hedges according to stock price. Remark that the purpose of those figures is not to present the strongest hedging agents but to prove that after a relatively short training process MCTS already acquired the essential hedging information. Notice also that we cannot run this comparison in the HN example, as switching off transaction costs yields a utility function that has no maximum. In all examples the introduction of transaction costs has no apparent impact on the convergence of the training process if collected costs are added as part of the state variable. Stochastic volatility, to the contrary, has a significant impact because much more training samples are required for the agent to learn the distribution of variance.

For comparison we also implemented a plain DQN hedger for terminal utility maximization, where the architecture was chosen such that all reward is granted at maturity. However, training and runtime performance were very poor. The reason lies in the fact that reward is granted only at maturity. DQN must identify those sequences of actions that yield positive rewards from an exponentially growing set of admissible action sequences. Given the $21^{20}$ possible decision paths our examples seem out of reach for plain Monte Carlo and tabula rasa DQN. Notice that this implementation differs from previous successful reports, e.g.~\cite{kolm2019dynamic,cao2019deep} in that the agent receives no reward signal before maturity.
\section{Conclusion}
We have introduced a new class of algorithms to the problem of hedging and pricing of financial derivative contracts. Our approach is based on state-of-the-art MCTS planning methods, whose architecture is inspired by the most successful systems for game-tree search. This is justified by the theoretical insight that discrete pricing and hedging models can be represented in terms of decision trees. As compared to DQN, MCTS combines RL with {search}, which results in a stronger overall performance. We report that our implementation of DQN maximization of terminal utility was not successful, while MCTS shows solidly improving reward curves within the computational resources available to us. We finally note that plain RL agents (including DQN-based agents) trained on simulated data will
perform well in market situations that are similar to the training simulations. This shortcoming is addressed by MCTS within the NN's generalization ability by evaluating multiple market states and active search. We leave the valuation of complex derivatives, including American-type or path dependent derivatives for further research, but it is conceivable that with sufficient resources granted, trained MCTS agents will show very competitive performance.
\clearpage
\bibliographystyle{plain}
\bibliography{Bibliography.bib}
\end{document}